\documentclass{article}
\usepackage{spconf,amsmath,amssymb,graphicx,booktabs,hyperref,xcolor}

\title{Lie to me: Detecting Managerial Evasiveness in 
\\ Earnings Calls via Conversational Audio Encoders}

\name{Huizhong Chen$^{\star}$ \qquad Huan Zhang$^{\dagger}$}
\address{$^{\star}$Beijing Normal-Hong Kong Baptist University $^{\dagger}$Individual researcher \\
\texttt{u430018007@mail.bnbu.edu.cn}}

\begin{document}
\maketitle

\begin{abstract}
Earnings conference calls are a primary channel through which managers disclose information under analyst scrutiny \cite{frankel1999empirical}. Prior work has linked vocal and lexical cues to future adverse outcomes, but often pools features over an entire call
and underuses the interactive structure of Q\&A.
We propose a two-branch late-fusion framework for detecting \emph{managerial evasiveness} as a predictor of
extrinsic SEC events (primarily late filings):
(i)~an \textbf{LLM-as-a-judge} that maps Q\&A text to an interpretable call-level vector $X_{\text{text}}$ via a
structured binary rubric, and
(ii)~a \textbf{frozen conversational encoder} whose temporal hidden states are read by a
DeepVoice-style sequential reader to produce an audio representation $h$.
Late fusion of $(X_{\text{text}},h)$ yields a call-level risk score $p$.
On $n{=}1{,}039$ calls ($212$ late filings) with firm-grouped 5-fold CV, fusion reaches AUROC $\approx 0.89$, versus $0.55$ for the text judge and $0.71$ for duration alone.
These results show that conversational audio dynamics encode managerial evasiveness beyond lexical content and call length, yielding a stronger early-warning signal of adverse SEC outcomes.
\end{abstract}


\section{Introduction}
\label{sec:intro}

When firms face adverse fundamentals, managers in earnings-call Q\&A often avoid direct answers:
they pivot to strategy, deflect to colleagues, hedge numbers, or stall before responding.
Such \emph{evasiveness} is not identical to financial-statement fraud, but it is a behaviorally
interpretable signal that may precede restatements, delayed filings, or distress.

Classic accounting studies extract vocal affect or linguistic deception cues from calls~\cite{frankel1999empirical, Larcker2012, Baik2025}, and recent multimodal models fuse verbal and vocal streams for fraud detection~\cite{lu2025cmmd}.
Most pipelines, however, (a) pool features across the whole call~\cite{mayew2012power}, (b) treat Q\&A turns as independent snippets~\cite{Gow2021}, or (c) use single-stream audio encoders that were not trained for dialogue~\cite{ewertz2026listen}.

We argue that evasiveness is fundamentally \emph{interactive}: it depends on what was asked,
how hard it was pressed, and how the manager times and structures the reply within an ongoing dialogue~\cite{nuaimi2025detecting}.
This paper instantiates a compact late-fusion architecture (Fig.~\ref{fig:arch}):

\begin{itemize}
\item \textbf{LLM-as-a-judge (text).}
Following the LLM-as-a-judge paradigm \cite{Zheng2023Judge}, an LLM applies a fixed binary rubric to
analyst--manager exchanges and aggregates tags into $X_{\text{text}}$.
\item \textbf{Frozen Moshi with a sequential audio head.}
We extract temporal Moshi hidden states from Q\&A audio \cite{Defossez2024Moshi} and train only a
lightweight DeepVoice-style sequential reader \cite{Yang2023DeepVoice}, keeping the
dialogue foundation model frozen.
\end{itemize}

We evaluate this architecture on $1{,}039$ earnings calls linked to their subsequent SEC filings, and we
make three contributions.
First, we assemble a call-level resource that pairs publicly posted Q\&A audio and captions with EDGAR
late-filing and restatement records, so that supervision is extrinsic to the call and is never observed by
the judge.
Second, we show that a frozen dialogue foundation model, read by a lightweight sequential head, predicts
late filings far more accurately than the interpretable text rubric or simple call-length controls, and
that this ordering persists when both modalities are given the same classifier family.
Third, we establish that the audio signal is incremental and call-specific: it remains informative after
controlling for $X_{\text{text}}$, and it disappears when audio is reassigned across calls, which rules out a
corpus-wide prior and locates the signal in the delivery of the individual exchange.

\section{Related Work}
\label{sec:related}

\subsection{Vocal and delivery cues in dialogue}

Foundational work in accounting has established that vocal cues carry predictive signal, with studies linking affective acoustic features to future firm performance and vocal delivery quality to information processing costs \cite{mayew2012power,Baik2025}—yet these studies typically aggregate over the manager's entire speech rather than modeling Q-A contingencies~\cite{hollander2010silence}. Concurrently, lexical deception markers and explicitly coded non-answers demonstrate that text alone can flag misleading or evasive language \cite{Larcker2012,Gow2021}. However, these text-based detectors capture either overt linguistic patterns or clear-cut refusals to answer, leaving undetected the broader and more subtle class of evasive responses that provide surface-level answers while deflecting substance~\cite{waddle2020important}. The most adjacent prior work is a contrastive multimodal dialogue network (CMMD) that fuses verbal and vocal streams across analyst-manager turns for financial fraud detection. Yet that model is an end-to-end heavy system~\cite{lu2025cmmd} (comprising four modules including multimodal fusion with co-attention and contrastive learning, and dialogue interaction learning, trained on RTX 3090 with 10-fold cross-validation) optimized for rare fraud labels~\cite{branco2016survey}, whereas our setting targets evasiveness as a behaviorally interpretable construct and uses a lighter two-branch pipeline.
The unit of analysis is the Q\&A exchange, the target is evasiveness rather than fraud, and the method pairs an LLM judge with a frozen conversational encoder so each branch can be ablated.

\subsection{Audio LLMs and duplex dialogue models}

General-purpose audio LLMs like GPT-4o~\cite{openai2024gpt4o} enable zero-shot end-to-end reasoning over Q\&A audio, offering a natural upper-bound baseline for prompt-only evasiveness scoring, yet their opaque and dyad-unaware representations do not expose stable, human-interpretable feature groups for targeted ablation~\cite{lipton2018mythos}.
Utterance-level self-supervised encoders such as WavLM and emotion2vec yield strong content or affect embeddings, but they remain single-stream and are typically pooled over a speaker's span---the same aggregation used in vocal-affect studies of earnings calls~\cite{Chen2022WavLM,Ma2024emotion2vec}.
Dual-channel spoken-dialogue models such as dGSLM show that gaps, overlaps, and role timing can be learned from raw conversational audio~\cite{Nguyen2023dGSLM}, which is the representation we want for Q\&A evasiveness.
Full-duplex models such as Moshi are built for that setting: their dual-stream transformer processes speech-to-speech dynamics, while the Mimi tokenizer compresses audio into representations that encode response gaps, overlaps, and role asymmetry, precisely the conversational contingencies (what was asked and how the manager timed the reply) that define evasiveness in Q\&A \cite{Defossez2024Moshi}.
These dialogue-structured hidden states make Moshi a natural fit for our acoustic branch, complementing the text-based LLM judge with turn-taking-aware features rather than aggregate vocal affect.
We isolate that contribution against the text judge in Table~\ref{tab:ablation}; prompt-only audio-LLM scoring is left as an orthogonal baseline for future work.

\section{Method}
\label{sec:method}

\begin{figure*}[t]
\centering
\includegraphics[width=\textwidth]{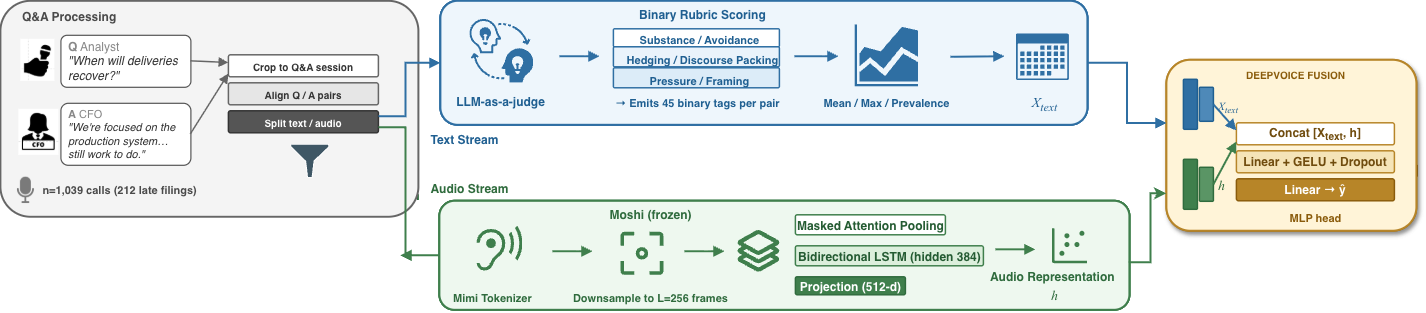}
\caption{Two-branch fusion architecture.
Prep yields Q\&A text/audio.
An LLM-as-a-judge produces $X_{\text{text}}$; a frozen Moshi encoder yields frame states read by a DeepVoice-style reader to $h$;
late fusion maps $(X_{\text{text}},h)$ to $p$ for extrinsic SEC label $y$.}
\label{fig:arch}
\end{figure*}

\subsection{Data}
\label{sec:data}

We study publicly posted earnings calls from AlphaStreet's YouTube channel \cite{AlphaStreetYouTube},
catalogued with video identifiers, tickers, fiscal period, and duration.
Related multimodal earnings-call corpora such as MAEC \cite{Li2020MAEC} motivate the pairing of text with
audio; we assemble our sample independently from AlphaStreet and EDGAR rather than redistributing MAEC.
Tickers are mapped to SEC Central Index Keys (CIKs), unique identifiers assigned to SEC filers, to link each company with its corresponding EDGAR filings
($4{,}278$ explicit-ticker, $1{,}047$ exact-name, and $284$ partial-name matches, with $1{,}032$ unmatched),
and unmatched firms cannot receive filing-based positives.
We estimate each call date as fiscal-quarter end plus 35~days and then search EDGAR over the following
24~months.
The primary label $y{=}1$ is an NT~10-K/NT~10-Q \textbf{late-filing} notice under Rule~12b-25
\cite{SECRule12b25,Bartov2017Late} ($223$ calls);
8-K Item~4.02 \textbf{restatements} ($66$) form a second outcome, and we combine the two into a broader
adverse-event label that is positive whenever a call is followed by either event ($252$ calls). Both outcomes capture adverse post-call developments, with late filings indicating reporting or compliance difficulties and 8-K Item 4.02 events indicating that previously issued financial statements should no longer be relied upon.

We also retrieve captions for every adverse-event call, which yields
$1{,}101$ caption-ready calls out of $6{,}641$ catalogued.
Fusion additionally requires Q\&A audio, a judge output $X_{\text{text}}$, and conversational audio encoder frames in
$\mathbb{R}^{T\times 4096}$, and the intersection of these three modalities defines the analysis sample of
$n{=}1{,}039$ calls ($212$ late and $827$ non-late) drawn from $548$ distinct firms.

\subsection{Problem setup}
\label{ssec:setup}

For each call $i$ in this sample we observe Q\&A text and Q\&A audio, and we predict the extrinsic label
$y_i\in\{0,1\}$ introduced above, primarily an NT~10-K/NT~10-Q late-filing notification within 24~months of
the estimated call date \cite{Bartov2017Late,SECRule12b25}, with the restatement and adverse-event labels
as secondary outcomes.
Late filings are economically consequential in capital markets \cite{Bartov2017Late}, yet they are not
identical to fraud; we treat them as a noisy but auditable risk label.
We report firm-grouped CV AUROC and test (A)~whether text$+$audio fusion beats text alone and
(B)~whether audio still helps after controlling for $X_{\text{text}}$.

We parse the auto-captions into Q\&A, timestamps, and optional pair spans from operator/analyst cues.
When a Q\&A window is detected we restrict features to that span; otherwise we use the available continuous Q\&A text/audio.

\subsection{Text branch: LLM-as-a-judge }
\label{ssec:llm}

Accounting work shows that explicit non-answers and lexical deception markers are informative
\cite{Larcker2012,Gow2021}.
We turn Q\&A text into $X_{\text{text}}$ with a frozen LLM-as-a-judge \cite{Zheng2023Judge}.
Captions are split into analyst--manager exchanges at operator and analyst cues.
We drop operator boilerplate and spans shorter than 30 words, then keep at most the first eight remaining
exchanges.
Moonshot \texttt{kimi-k2.6} labels each exchange independently with a frozen checklist of $45$ yes/no tags,
grouped as in Fig.~\ref{fig:arch} into substance versus avoidance, hedging and discourse packing, and
pressure and framing.
Substance and avoidance tags ask whether the manager answered the question that was asked with a number,
range, timeline, or a clear yes/no, or instead refused, took the item offline, or claimed they could not
disclose; hedging and packing tags mark hedges, process talk, long sidesteps, and safe-harbor boilerplate;
pressure and framing tags mark whether the ask was clear, whether the manager spoke to it, and whether the
reply reframed the metric or blamed only exogenous factors.
Every tag must be $0$ or $1$, with a short rationale that cites wording in the exchange; the model is not
asked to score risk.
The judge never sees filings, returns, or $y$; a missing tag is $0$.

A call with $n\le 8$ scored exchanges is therefore an $n\times 45$ matrix of bits.
For each tag we keep its mean and its maximum over the $n$ exchanges, and the dump stores the mean a
second time under another name.
We also keep $n$, and two call-level summaries of the same tags: how often the manager gave a checkable
answer, how often they refused or deflected, and the difference of those two rates.
That is $45\times 3+1+3=139$ floats, read in sorted key order by every model that uses $X_{\text{text}}$.
The only later transform is fold-wise standardization.

\subsection{Audio branch: Interactive utterances}
\label{ssec:moshi}

Prior vocal studies in earnings calls typically pool affect or delivery features over long spans
\cite{mayew2012power,Baik2025,Yang2023DeepVoice}.
Our setting is interactive Q\&A, so we want an encoder pretrained for \emph{dialogue} rather than
isolated utterances.
\textbf{Moshi} is a full-duplex speech--text foundation model whose Mimi codec and temporal transformer
are trained for real-time conversational turn-taking \cite{Defossez2024Moshi}.
Concretely, Q\&A audio (24\,kHz) is mapped to Mimi tokens and, through the released \texttt{moshiko}
checkpoint, to hidden states $F\in\mathbb{R}^{T\times 4096}$.
Analyst and manager audio enter as two separate Mimi code streams, so the encoder receives the exchange as
a dialogue rather than as one undifferentiated signal.
We take $F$ from the last layer of the $32$-layer temporal transformer, the state from which Moshi predicts
the next token \cite{Defossez2024Moshi}; its width of $4096$ is the transformer hidden size, and its rate of
$12.5$\,Hz is the Mimi frame rate.
Sequences are uniformly subsampled to at most $L{=}256$ frames, and the encoder itself is never updated.

On top of these frozen frames we train a DeepVoice-style sequential reader, following the vocal risk
classifier of Yang et al.\ \cite{Yang2023DeepVoice}.
For frame $t$, a learned projection with GELU \cite{hendrycks2016gelu} and dropout maps
$F_t\in\mathbb{R}^{4096}$ to $z_t\in\mathbb{R}^{512}$.
A one-layer bidirectional LSTM with 384 units in each direction produces
$q_t\in\mathbb{R}^{768}$.
Masked attention assigns
$\alpha_t=\operatorname{softmax}_t(w^\top q_t+b)$ and forms the call representation
$h=\sum_t\alpha_tq_t$, so the reader can weight frames rather than averaging the call.
The audio probability
$\hat p_{\text{audio}}=\sigma(g_{\text{audio}}(h))$ is this branch's score that the call is followed by a late
filing, where $\sigma$ is the logistic function and
$g_{\text{audio}}$ is a $768{\rightarrow}384{\rightarrow}1$ nonlinear readout with GELU and dropout.
Only these three components, projection, recurrent layer, and readout, carry gradients, which separates a
dialogue-pretrained representation from a small supervised model of the task,
in contrast to end-to-end multimodal fraud detectors such as CMMD \cite{lu2025cmmd}.

\subsection{Late fusion}
\label{ssec:fusion}

Within each fold we standardize the frozen judge vector $X_{\text{text}}\in\mathbb{R}^{139}$, concatenate it
with the reader's $h\in\mathbb{R}^{768}$, and classify $[X_{\text{text}}\|h]\in\mathbb{R}^{907}$ with a
$907{\rightarrow}453{\rightarrow}1$ head that uses GELU and dropout, matching the audio readout
(Fig.~\ref{fig:arch}).
Moshi remains frozen; the projection, LSTM, attention, and fusion head are trained jointly on the
late-filing loss.
A class-balanced logistic regression on $[X_{\text{text}}\|\hat p_{\text{audio}}]$ collapses audio to the
scalar from the audio-only branch; we report its $\Delta$AUROC against the otherwise identical text-only
logistic regression.

\section{Experiments}
\label{sec:exp}

\begin{table*}[t]
\begin{minipage}[t]{0.48\linewidth}
\centering
\caption{Late-filing prediction, firm-grouped 5-fold CV, $n{=}1{,}039$ ($212$ positives).}
\label{tab:ablation}
\small
\begin{tabular}{@{}lcc@{}}
\toprule
Predictor & AUROC & $\Delta$ vs.\ text LR \\
\midrule
Text rubric $+$ logistic & $0.553$ & -- \\
Text rubric $+$ nonlinear head & $0.520$ & $-0.033$ \\
Mean Moshi $+$ logistic & $0.818$ & $+0.265$ \\
Mean Moshi $+$ nonlinear head & $0.898$ & $+0.345$ \\
Moshi$+$DeepVoice-style reader & $0.902$ & $+0.349$ \\
Late fusion & $0.897$ & $+0.344$ \\
Text $+$ audio score & $0.851$ & $+0.298$ \\
\bottomrule
\end{tabular}
\end{minipage}\hfill
\begin{minipage}[t]{0.48\linewidth}
\centering
\caption{Ablations on the same $n{=}1{,}039$ sample.
Reference (seed $0$, $L{=}256$, 40 epochs): audio/fusion $0.902$/$0.897$.}
\label{tab:robust}
\small
\setlength{\tabcolsep}{4pt}
\begin{tabular}{@{}lcc@{}}
\toprule
Variant & Audio & Fusion \\
\midrule
\multicolumn{3}{@{}l}{\textit{Sensitivity to label and configuration}} \\
Multi-seeds average & $0.847{\pm}0.023$ & $0.893{\pm}0.002$ \\
Late filing or restatement & $0.895$ & $0.893$ \\
\addlinespace
\multicolumn{3}{@{}l}{\textit{Perturbed audio input}} \\
Frames shuffled within call & $0.830$ & -- \\
Frames reassigned across calls & $0.483$ & -- \\
\bottomrule
\end{tabular}
\end{minipage}
\end{table*}

\subsection{Setup}
\label{ssec:setupexp}

On the $n{=}1{,}039$ intersection we run the models in Table~\ref{tab:ablation} under five-fold
CV grouped by company (SEC CIK; seed~$0$ is the designated run; further seeds in
Sec.~\ref{ssec:robust}).
The main audio model is the DeepVoice-style reader of Sec.~\ref{ssec:moshi} on frozen Moshi frames
($L{=}256$).
Late fusion concatenates $X_{\text{text}}$ with that reader's $h$; we also put
$\hat p_{\text{audio}}$ beside $X_{\text{text}}$ in a logistic regression, as in Sec.~\ref{ssec:fusion}.
To keep modality separate from classifier family we run the same two heads on the text rubric and on
mean-pooled Moshi frames $\bar F=T^{-1}\sum_t F_t$: class-balanced logistic regression, and a
one-hidden-layer nonlinear head (128 units, GELU, dropout $0.1$).
Duration and Q\&A word count are length controls.

The metric is mean fold AUROC.
Logistic models use balanced class weights.
Neural heads use positive-class-weighted binary cross-entropy, AdamW
($3{\times}10^{-4}$, weight decay $10^{-4}$), gradient clipping at 1, and no Moshi fine-tuning.
The DeepVoice and fusion heads use batch size 8, dropout $0.1$, and 40 epochs;
the matched nonlinear heads use batch size 32 and 40 epochs.
Standardization is fitted on training calls in each fold.

\subsection{Main result}
\label{ssec:mainres}

Table~\ref{tab:ablation} reports seed-$0$ firm-grouped AUROC on the same $n{=}1{,}039$ calls.
The text rubric is a weak ranker of late filings: logistic regression reaches $0.553$, and giving it the
same nonlinear head used on audio does not help ($0.520$).
With those same two heads, mean-pooled Moshi frames reach $0.818$ and $0.898$.
Audio uses the extra capacity; the tags do not, so the gap is the representation rather than the
classifier family: frozen conversational frames already outperform the judge vector before any LSTM is
applied.
The DeepVoice-style reader adds only a small increment over mean pooling ($0.902$ versus $0.898$), so most
of the audio signal is in the Moshi frames rather than in sequential modeling.

Late fusion of $X_{\text{text}}$ with the reader's $h$ reaches $0.897$, matching the audio-only reader
rather than beating it.
Concatenating the rubric into the head therefore does not improve ranking once audio is present.
Putting $\hat p_{\text{audio}}$ beside $X_{\text{text}}$ in logistic regression collapses audio to one
score and still raises AUROC by $+0.298$ over text alone ($0.851$): below fusion, because $h$ is discarded,
but the increment remains after the rubric is controlled.
Seed-$0$ audio at $0.902$ is the high draw; variation across seeds is in Sec.~\ref{ssec:robust}.

\subsection{Ablations}
\label{ssec:robust}

Table~\ref{tab:robust} checks whether the audio advantage depends on the label and the random seed,
and what the audio branch is actually reading.
Replacing late filings with the adverse-event label of Sec.~\ref{sec:data}---positive if a late filing
\emph{or} an 8-K Item~4.02 restatement follows the call---gives $0.895$ for audio and $0.893$ for
fusion, so the ranking is not tied to one filing form.
Across three seeds, fusion is stable at $0.893{\pm}0.002$ while audio alone varies more
($0.847{\pm}0.023$).
Halving or quadrupling the frame budget, or training for 120 rather than 40 epochs, moves fusion by at most
$0.017$, so neither the temporal resolution nor extra optimization is carrying the result.

The lower block perturbs the audio input.
Shuffling the frame order within a call costs $0.072$ AUROC ($0.830$), below mean pooling of the original
frames ($0.898$): destroying order is not the same as averaging, yet most of the ranking survives.
Reassigning entire frame sequences to other calls is the decisive test: AUROC falls to $0.483$, which is
chance, and the increment over the rubric disappears.
The prediction thus depends on the acoustics of the specific call rather than on a corpus-wide prior that
any audio would satisfy.

\section{Conclusion}
\label{sec:conclusion}

We presented a two-branch fusion model of earnings-call Q\&A that pairs an interpretable
LLM-as-a-judge rubric with a frozen dialogue foundation model read by a lightweight sequential head.
Under firm-grouped cross-validation on calls linked to subsequent SEC filings, conversational audio is a
substantially stronger predictor of late filings than the text rubric or call length, reaching
AUROC $\approx 0.89$, and it remains informative once the rubric is controlled.
Because the signal falls to chance when audio is reassigned across calls, what the model reads is the
delivery of a particular exchange rather than a property of the corpus.
This suggests that how managers answer under analyst pressure carries disclosure-risk information that
lexical summaries of what they say do not, and that frozen dialogue encoders offer a practical way to
reach it.

\vfill\pagebreak
\clearpage
\bibliographystyle{IEEEbib}
\bibliography{strings,refs}

\end{document}